\documentclass[lettersize,journal]{IEEEtran}
\usepackage{amsmath,amsfonts}
\usepackage{algorithmic}
\usepackage{algorithm}
\usepackage{array}
\usepackage[caption=false,font=normalsize,labelfont=sf,textfont=sf]{subfig}
\usepackage{textcomp}
\usepackage{stfloats}
\usepackage{url}
\usepackage{verbatim}
\usepackage{graphicx}
\usepackage{cite}

\usepackage{booktabs,ragged2e}

\usepackage{threeparttable}

\usepackage[hidelinks]{hyperref}
\hypersetup{
    colorlinks=true,
    linkcolor=blue,
    filecolor=magenta,      
    urlcolor=blue,
    pdftitle={Overleaf Example},
    pdfpagemode=FullScreen,
    citecolor=black,
    }

\usepackage{breakurl}
\usepackage[breaklinks]{}

\usepackage{makecell} % Para quebrar células
\usepackage{tabularx} % Para controle automático de largura

\usepackage{booktabs,threeparttable}

\usepackage{tabularray}
\usepackage{arydshln}
\usepackage{booktabs}
\usepackage{multirow}
\usepackage[table,xcdraw]{xcolor}
\usepackage{arydshln}
\usepackage{graphicx}

\usepackage{orcidlink}

\usepackage{arydshln}

\usepackage[skip=1ex,font=small, labelfont=bf]{caption}

\begin{document}

% \title{CORTEX: A Simulation-Based Framework for Corner Case Testing and Exploration in Autonomous Vehicles }
\title{CORTEX-AVD: A Framework for CORner Case Testing and EXploration in Autonomous Vehicle Development}

\author{Gabriel Kenji Godoy Shimanuki \orcidlink{0000-0002-5412-8910}, Alexandre Moreira Nascimento \orcidlink{0000-0002-4342-6453}, Lucio Flavio Vismari \orcidlink{0000-0002-3101-2464}, 
João Batista Camargo, Jr. \orcidlink{0000-0001-5098-6769}, Jorge Rady de Almeida, Jr. \orcidlink{0000-0003-3839-4570}, Paulo Sergio Cugnasca \orcidlink{0000-0002-5675-4667}
        % <-this % stops a space
% \thanks{Submitted 09 April 2025.}
\thanks{Manuscript preprint resubmitted on April 09, 2025.}

\thanks{G. K. G. Shimanuki, A. M. Nascimento, L. F. Vismari, J. B. Camargo, Jr., and J. R. de Almeida, Jr., and P. S. Cugnasca, are with the Safety Analysis Group, Polytechnic School, University of São Paulo, São Paulo 05508-010, Brazil (e-mail: gabrielshimanuki@usp.br; alexandremoreiranascimento@alum.mit.edu}
}

% \markboth{IEEE TRANSACTIONS ON INTELLIGENT TRANSPORTATION SYSTEMS, VOL. XXX, NO. YYY, APRIL 2025}{IEEE TRANSACTIONS ON INTELLIGENT TRANSPORTATION SYSTEMS, VOL. XXX, NO. YYY, MARCH 2025}
\markboth{PRE-PRINT VERSION, APRIL 2025}{PRE-PRINT VERSION, APRIL 2025}

% \IEEEpubid{0000--0000/00\$00.00~\copyright~2025 IEEE}

\maketitle

\begin{abstract}

% Something here...

Autonomous Vehicles (AVs) aim to improve traffic safety and efficiency by reducing human error. However, ensuring AVs reliability and safety is a challenging task when rare, high-risk traffic scenarios are considered. These 'Corner Cases' (CC) scenarios, such as unexpected vehicle maneuvers or sudden pedestrian crossings, must be safely and reliably dealt by AVs during their operations. But they are hard to be efficiently generated. Traditional CC generation relies on costly and risky real-world data acquisition, limiting scalability, and slowing research and development progress. Simulation-based techniques also face challenges, as modeling diverse scenarios and capturing all possible CCs is complex and time-consuming. To address these limitations in CC generation, this research introduces \texttt{CORTEX-AVD}, an open-source framework that integrates the CARLA Simulator and Scenic to automatically generate CC from textual descriptions, increasing the diversity and automation of scenario modeling. Genetic Algorithms (GA) are used to optimize the scenario parameters in six case study scenarios, increasing the occurrence of high-risk events. Unlike previous methods, \texttt{CORTEX-AVD} incorporates a multi-factor fitness function that considers variables such as distance, time, speed, and collision occurrence. Additionally, the study provides a benchmark for comparing GA-based CC generation methods, contributing to a more standardized evaluation of synthetic data generation and scenario assessment. Experimental results demonstrate that the \texttt{CORTEX-AVD} framework significantly increases CC incidence while reducing the proportion of wasted simulations.

\end{abstract}


\section{Introduction}

    Artificial intelligence (AI), particularly machine learning (ML), is enabling new applications in several areas, including vehicle driving automation based on intelligent control algorithms, known as Autonomous Vehicle (AV) \cite{ma2020artificial, bathla2022autonomous}. The goals of the AV field include improving traffic safety and efficiency \cite{duarte2018impact, litman2020autonomous} by reducing accidents commonly associated with human error \cite{nascimento2019systematic, abdel2024matched, wang2020safety}.

    AVs are safety-critical systems, as failures can have serious consequences, including environmental damage, and financial or life losses \cite{nascimento2019systematic, naufal20172, chattopadhyay2020autonomous}. To mitigate AV safety risks, widely adopted industry standards, such as SAE J3016, which defines driving automation levels \cite{sae_av}, and ISO 26262, which outlines functional safety requirements for road vehicles \cite{iso201826262}, provide essential guidelines for the development of AV \cite{rajabli2020software}. These standards emphasize the need for rigorous validation processes to ensure AVs can handle a wide range of scenarios, including rare and unpredictable situations. Consequently, identifying and incorporating these challenging scenarios, often referred to as Corner Cases (CCs), is necessary for developing robust control algorithms capable of enhancing AV safety \cite{koopman2018toward, sun2021scenario, shimanuki2025navigating}.

    CCs represent atypical scenarios that rarely occur in everyday driving, but can lead to severe consequences if not handled properly. Examples include unexpected vehicle behavior, sudden pedestrian crossings, environmental factors that perturb sensors, or obstacles on the road \cite{koopman2016challenges}. However, some current approaches are heavily based on real-world data collection, which is costly, time-consuming, and inherently limited in capturing the diversity of rare events \cite{horel2022using, nascimento2019systematic, xu2021reliability, ge2022heterogeneous}. Physical tests, such as those performed at testing facilities like Mcity, located at the University of Michigan \cite{briefs2015mcity}, or Waymo's Castle \cite{cerf2018comprehensive, waymo_castle}, offer controlled environments but fail to cover the full spectrum of potential CCs \cite{koopman2016challenges}. To mitigate these issues, researchers are focusing on simulation-based methods to identify and generate CCs more efficiently \cite{kuutti2020survey, shimanuki2025navigating}.

    Simulated environments offer a promising alternative for controlled and repeatable AV testing in high-risk scenarios \cite{porsche2023, nvidia2023, gulino2024waymax, jiang2024scenediffuser, peng2024improving, mahjourian2024unigen}, yet generating diverse CCs remains a complex task \cite{koopman2016challenges}. Current methodologies often rely on labor-intensive scenario modeling, expert domain knowledge, or proprietary tools, limiting accessibility and slowing progress in developing robust control algorithms \cite{li2024choose, song2024industry, xu2022safebench, koopman2024redefining}. Furthermore, the black-box nature of Deep Learning (DL), the predominant model in AV decision-making \cite{nascimento2019systematic, kuutti2020survey, ma2020artificial}, complicates scenario validation \cite{koopman2016challenges, rajabli2020software}, making it difficult to ensure safe behavior under CC conditions \cite{norazman2023artificial, mechernene2022detection, kuutti2020survey}. These issues raise concerns about transparency and public trust in AV technology \cite{koopman2024anatomy, reuters2024tesla}. Although synthetic data generation and simulated environments have driven safety improvements, academic research remains fragmented, delayed by the absence of standardized benchmarks or unified testing frameworks, which limits meaningful collaboration between academia and industry \cite{li2024choose, song2024industry, xu2022safebench, koopman2024redefining}. Given the safety-critical nature of AVs, establishing open and standardized practices for CC generation is necessary to advance AV safety \cite{song2024industry, shimanuki2025navigating}. 

    To address these challenges, this study presents \texttt{CORTEX-AVD}, a high-level abstraction framework that integrates Carla Simulator and Scenic to identify CCs based on textual descriptions. By optimizing scenario modeling parameters, the framework increases the likelihood of generating CCs, thereby potentially enabling a more effective evaluation of AV safety and reliability performances. 
    
    In summary, this paper makes the following key contributions:

    \begin{itemize}
        \item Lightweight framework integrating Carla and Scenic to automatically generate CC from textual descriptions.
        \item Benchmark comparison of related studies using common metrics to evaluate Genetic Algorithm effectiveness.
        \item Comprehensive evaluating mechanisms for assessing traffic scenario metrics.
        \item A case study demonstrating improved risk scenario refinement and simulation validity.
    \end{itemize}

    This study is structured into 7 sections. Section \ref{sec:related-work} presents the related work. Then, Section \ref{sec:methodology} presents the methodology. Section \ref{sec:case_study} presents the proposed case study. Section \ref{sec:results} presents the results. Finally, Section \ref{sec:discussion} provides a discussion and Section \ref{sec:conclusions} the concluding remarks on the findings.  

\section{Related Work} \label{sec:related-work}

    Recent literature highlights the challenge of improving the robustness of control systems that rely on ML and DNN techniques, particularly through exhaustive and systematic testing \cite{tian2018deeptest, zohdinasab2021deephyperion, wang2020metamorphic}. This challenge is amplified by the fact that, while human drivers intuitively rely on prediction and reflexes to avoid accidents \cite{wang2021human}, AVs face the complex issue of systematically identifying and handling risky scenarios during their operation. As a result, much of the research focuses on developing methods to identify and generate CCs where AVs are likely to fail \cite{sun2021corner}, as these scenarios provide valuable test data for evaluating AV performance under diverse and challenging conditions \cite{bolte2019towards}.

    A common approach for generating CC employ reinforcement learning, frequently integrated with adversarial or generative methods \cite{gupta2022towards, niu2021dr2l, vardhan2021rare, du2023learning, du2021adaptive, karunakaran2020efficient, niu2023re, kang2023ecsas, hao2023adversarial, dagdanov2023self, zhu2023rita}. For example, reinforcement learning combined with adversarial techniques is commonly used to create hostile driving environments \cite{gupta2022towards, du2021adaptive, niu2023re, karunakaran2020efficient}. Deep reinforcement learning methods are also applied to generate CCs, allowing agents to learn from simulated environments and develop policies for rare, high-risk events \cite{niu2023re, dagdanov2023self}. However, these methods face several limitations. The PAIN framework, for example, is constrained by a limited field-of-view, narrowing its ability to model dynamic environments like rear-end collisions \cite{gupta2022towards}. Similarly, while the DR2L method offers valuable insights, it fails to account for real-world complexities such as varying weather conditions and difficult road geometries, leaving real-world validation as a significant gap \cite{niu2021dr2l}. The RARE framework can identify CCs, but the high computational costs of extensive scenario testing limit its practical scalability \cite{vardhan2021rare}. Further, generative adversarial networks and reinforcement learning methods, while promising, face scalability issues that limit their ability to replicate diverse conditions \cite{du2023learning, karunakaran2020efficient}. Methods relying on narrow datasets or synthetic data can overfit, which affects their generalizability in new environments \cite{du2021adaptive}. In addition, some methods are overly focused on specific accident types, reducing their applicability to broader driving contexts \cite{niu2023re}. Finally, the RITA framework struggles to replicate human behavior, such as unpredictable driver actions or pedestrian intentions, limiting its ability to identify real-world CCs \cite{zhu2023rita}. Consequently, addressing these limitations calls for exploring alternative approaches capable of enhancing diversity, scalability, and realism in CC generation.

    A notable category of algorithms for generating CC is evolutionary search methods, with Genetic Algorithm (GA) being the most frequently discussed. GA works by evolving and refining test scenarios through the selection, combination, and mutation of a set of cases to create new but more critical ones. Numerous studies using GA highlight their success in generating CCs \cite{li2020av, luo2021targeting, langford2019applying, tang2021systematic, kluck2019genetic, kaufmann2021critical, zhou2023specification, kluck2023empirical, kluck2019performance, gambi2019automatically, humeniuk2023ambiegen, tian2022mosat, birchler2023single, abdessalem2018testing, ebadi2021efficient}. These algorithms are particularly effective in identifying rare and extreme situations, such as unusual collisions or unexpected vehicle interactions, which methods such as random search are less likely to detect.    
    
    Although recent studies highlight the effectiveness of GA in CC generation, most experiments are limited to a few thousand simulations and rely on single-objective functions based on narrow metrics such as time or distance  \cite{kluck2019performance, tian2022mosat, gambi2019automatically, birchler2023single, langford2019applying, tang2021systematic, ebadi2021efficient, kluck2019genetic, kaufmann2021critical, zhou2023specification, kluck2023empirical}. These approaches may neglect variations of CCs, as they may fail to adequately capture the complexity of real-world driving scenarios. By contrast, in other fields, multi-objective optimization enables a more comprehensive search, offering a broader exploration of potential solutions \cite{deb2011multi, sharma2022comprehensive}. However, in the context of AV CC generation, only a few studies using GAs rely on multi-objective optimization techniques \cite{tian2022mosat, birchler2023single, abdessalem2018testing, ebadi2021efficient}, allowing for the balancing of factors like safety, efficiency, and complexity.
    
    Besides GA, other evolutionary methods like novelty search and a broad many-objective optimization explore test scenarios \cite{zohdinasab2021deephyperion, langford2019applying}. Novelty search maximizes diversity, revealing neglected CCs, while MAP-Elites \cite{zohdinasab2021deephyperion} partitions the search space to uncover rare cases.
    
   Beyond the limitations previously discussed, these methods share common challenges. Techniques like \cite{zohdinasab2021deephyperion, li2020av, tang2021systematic, humeniuk2023ambiegen} face high computational demands, requiring specialized hardware and limiting scalability in complex scenarios. Some of these approaches struggle with simulation determinism and the realism of agent behavior \cite{abdessalem2018testing, kaufmann2021critical}. Thus, to address these challenges and enhance the applicability of these techniques, further research is needed to improve computational efficiency, scalability, and realism, with multi-factor approaches and standardized evaluation criteria to enable fair comparisons across studies and assess performance consistently.

\section{The \textbf{\texttt{CORTEX-AVD}} Framework}\label{sec:methodology}

    This section presents the proposed framework for automatically generating CCs to support the development of robust vehicle control systems based on high-risk driving scenarios. The following subsections describe the simulation infrastructure (\ref{sub-sec:simulation_infrastructure}), the framework used to implement GA (\ref{sub-sec:parameters_refinement}), and the metrics used to evaluate and compare experiments (\ref{sub-sec:evaluation_metrics}).

    % This section presents the proposed framework for automatically generating CCs to support the development of robust vehicle control systems based on high-risk driving scenarios. The following subsections describe the simulation infrastructure (\ref{sub-sec:simulation_infrastructure}), the framework used to implement GA (\ref{sub-sec:parameters_refinement}), the experimental design employed in the case study (\ref{sub-sec:experimental_design}), and the metrics used to evaluate and compare experiments (\ref{sub-sec:evaluation_metrics}).

    \subsection{Simulation Infrastructure} \label{sub-sec:simulation_infrastructure}

        The infrastructure used in this study was conceived by integrating the Carla Simulator and the Scenic programming language, aiming a robust platform for generating and testing AV scenarios. The Carla Simulator is an open-source platform developed at the Computer Vision Center of the Universitat Autonoma de Barcelona \cite{Dosovitskiy17}. It offers a realistic environment for simulating vehicle traffic scenarios and implementing control algorithms for AVs. Carla supports high-fidelity rendering, sensor simulation, and complex dynamic environments, making it the ideal open-source simulator for testing AV in a variety of driving conditions \cite{li2024choose}.

        The Scenic is a probabilistic programming language to define complex traffic scenarios by specifying spatial and temporal relationships between agents and physical entities, with constraints that can range from strict to more relaxed conditions \cite{fremont2023scenic}. Its concise syntax simplifies the generation of diverse and realistic driving scenarios \cite{li2024choose}. Specifically, Scenic was selected for its efficiency in generating scenarios from formal descriptions within Carla simulator \cite{fremont2023scenic}. However, the framework is not limited to Scenic and can be adapted to support other high-level, textual scenario description languages.

\begin{figure}[h]
            \centering
            \includegraphics[scale=0.39]{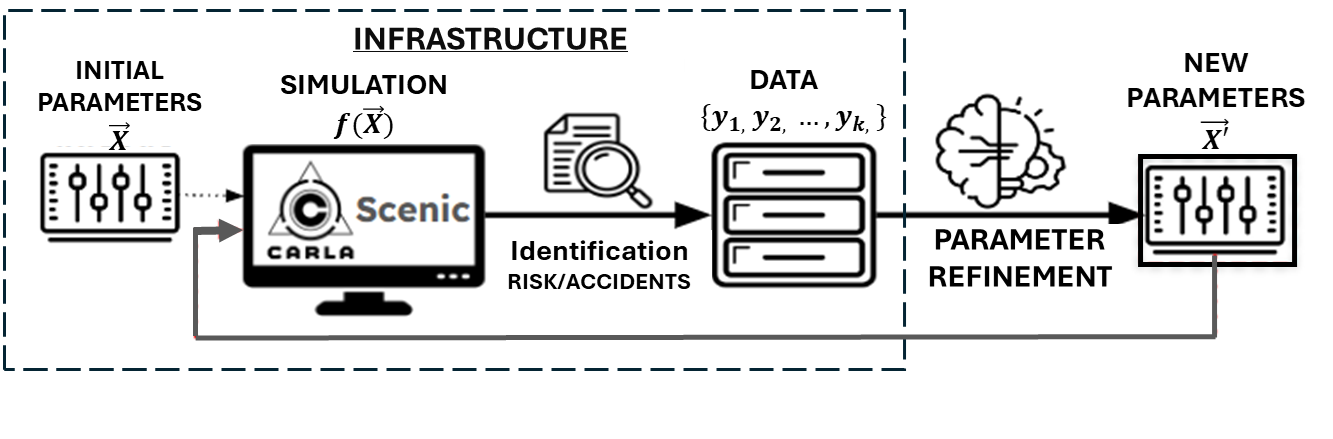}
            \caption{Simulation Infrastructure}
            \label{fig:simulation_infrastructure}
        \end{figure}

        Carla (0.9.13) and Scenic (2.1.0) were integrated on an Ubuntu 22.04 system with Python 3.8.10. As shown on the left side of Figure~\ref{fig:simulation_infrastructure}, this setup enabled the generation of simulation scenarios based on parameter vectors \( \vec{x} \in \mathbb{R}^n \). Each \( \vec{x} \) defines a set of simulation outputs \( f(\vec{x}) = \{ y_1, y_2, \dots, y_k \} \), where each \( y_i \in \mathcal{Y} \) represents an individual simulation instance. This system can be formally described as a function  $f: \mathbb{R}^n \rightarrow \mathcal{P}(\mathcal{Y})$, mapping input parameters to sets of data simulations. The resulting infrastructure supports efficient and scalable generation of AV testing scenarios, offering a robust platform for system validation.

    \subsection{Parameters Refinement} \label{sub-sec:parameters_refinement}

        In the framework, a Genetic Algorithm (GA) is employed to optimize the search for parameter vectors \( \vec{x} \in \mathbb{R}^n \) within Scenic scripts, aiming to generate high-risk scenarios such as accidents and near-misses. Inspired by natural selection, GAs have proven effective in navigating complex, nonlinear parameter spaces to identify critical conditions in dynamic systems \cite{deb2011multi, sharma2022comprehensive}. Traditional methods, such as manually mapping or exhaustively listing potential accident scenarios, are impractical due to the system’s high dimensionality and nonlinear behavior. In contrast, GAs are particularly well-suited for exploring large search spaces without relying on gradient information, avoiding costly derivative computations and handling non-differentiable or noisy objective functions with greater robustness \cite{forrest1993genetic}. Their scalability and global search capabilities enable efficient convergence to high-risk configurations, even in large-scale simulations. Moreover, GAs adapt seamlessly as new elements are introduced during scenario evolution. As shown on the right side of Figure~\ref{fig:simulation_infrastructure}, this optimization process forms a feedback loop where the GA adjusts the input parameters \( \vec{x} \), resulting in updated simulation outcomes \( f(\vec{x'}) = \{ y_1', y_2', \dots, y_k' \} \). This loop illustrates how the framework dynamically refines scenario configurations to uncover edge cases and critical testing conditions.

        \subsubsection{Genome Encoding Strategy}

            Scenic scripts use numeric parameter values and, for this reason, the genetic sequence in the GA was encoded as a vector of numeric values: \texttt{<EGO\_INIT\_DIST, EGO\_SPEED, EGO\_BRAKE, ADV\_INIT\_DIST, ADV\_SPEED, SAFETY\_DIST, CRASH\_DIST>}. Each parameter was assigned a continuous range based on a combination of preliminary experiments, empirical observations, and typical driving values found in simulation environments. Specifically, \texttt{EGO\_SPEED} and \texttt{ADV\_SPEED} were constrained to realistic urban and highway speeds $[5, 80]$ (km/h), while braking and safety-related parameters - \texttt{EGO\_BRAKE} $\in [0, 1]$, \texttt{SAFETY\_DIST} $\in [0, 20]$ (m), and \texttt{CRASH\_DIST} $\in [0, 5]$ (m) - were calibrated based on the behavior of the vehicle dynamics and threshold tuning across test runs. Initial distances for ego and adversary vehicles ($[0, +\infty[$) were empirically constrained during simulations to maintain feasibility and ensure timely interactions. This encoding strategy effectively captured the scenario’s variability while keeping the search space manageable and reproducible.

        \subsubsection{Fitness Function}

            By combining insights from the literature with the data collected from the vehicles during the simulations, hypotheses were formed regarding metrics useful for the objective function (Table \ref{tab:scenic_variables}) \cite{nascimento2019systematic, kim2022drivefuzz, zhu2023rita, shu2021test, tian2022generating, song2022scenario}. Based on these references, the multi-factor objective function was constructed as the sum of variables frequently associated with driving risk: Collision (C), Minimum Distance between vehicles (MD), Distance at Maximum Approach Speed (D\_MS), and Time-to-Collision (TTC) at the moment of Maximum Approach Speed (TTC\_MS) (Table \ref{tab:funcao_objetivo}). The function returns a value in the range $[0, 22]$, where higher scores indicate greater scenario risk. The function was evaluated through experiments to verify its ability to distinguish scenarios with different risk levels, with results presented in Section \ref{sec:results} indicating that it contributed meaningfully to the search process.
            
            In this formulation, a risk level of 12 can result from different combinations of scores. For example, a \texttt{Collision Occurrence} score of 10 with a combined score of 2 from the remaining variables, or equal scores of 4 across \texttt{MD}, \texttt{D\_MS}, and \texttt{TTC\_MS}. However, a risk score of 14 necessarily indicates a collision occurrence, since non-collision scenarios cannot exceed a total of 12.
            
            Additionally, due to the Scenic's sampling nature, some parameter sets may produce invalid or non-executable test cases. These are assigned a risk score of -1 and excluded from the GA process. During scenario generation, only valid cases are retained for selection, crossover, and mutation, ensuring that the optimization operates solely on executable simulations.

            \setlength{\tabcolsep}{1pt}
% \begin{table*}[h]
\begin{table}[h]
\caption{Scenic parameters}
\label{tab:scenic_variables}
\centering
% \begin{tabularx}{\textwidth}{|m{2.5cm}|m{9cm}|m{5cm}|}
% \begin{tabularx}{\textwidth}{|m{2.5cm}|m{2.8cm}|m{2.8cm}|}
\begin{tabularx}{\textwidth}{|m{1.5cm}|m{7cm}|}
 % \toprule
 \cmidrule(){1-2}
 \textbf{Type} & \textbf{Description} \\ 
 % \hline
 \cmidrule(){1-2}
 Event & \textbf{\underline{Collision occurrence}} \\
 % \hline
 \cmidrule(){1-2}
 \multirow{2}{*}{\scriptsize{Command}}      & Steering wheel oscillation \\
 & 
 Oscillation between pedals (accelerator, brake) \\
 % \hline
 \cmidrule(){1-2}

 \multirow{6}{*}{\scriptsize{Dynamics}} &

 \textbf{\underline{Minimum relative Distance between vehicles (MD)}} \\
 &
 Relative speed (approach) of vehicles at the instant of MD \\
 &
 Time-to-collision (TTC) of vehicles at the instant of MD \\
 &
 \textbf{\underline{Vehicles distance at Maximum Speed (MS)}} \\
 &
 Relative speed (approach) of vehicles at the instant of MS \\
 &
 \textbf{\underline{TTC of vehicles at the instant of MS}} \\
 % \hline
 \cmidrule(){1-2}
\end{tabularx}
% \end{table*}
\end{table}

            \begin{table}[h]
\caption{Risk associated with each metric}
\label{tab:funcao_objetivo}
\centering
% \begin{tabular}{@{}ccc@{}}
\begin{tabular}{@{}clc@{}}
\toprule
\scriptsize{\textbf{Metric}}         & \scriptsize{\textbf{Range}}    & \scriptsize{\textbf{Risk Score}} \\ \midrule
\multirow{2}{*}{\scriptsize{C}}       & \texttt{\scriptsize{True}}            & \scriptsize{10}                      \\
                         & \texttt{\scriptsize{False}}                & \scriptsize{0}                       \\ \hdashline
\multirow{5}{*}{\scriptsize{MD}}      & \scriptsize{${[}0, 820{[}$}          & \scriptsize{4}                       \\
                         & \scriptsize{${[}820, 1100{[}$}       & \scriptsize{3}                       \\
                         & \scriptsize{${[}1100, 1376{[}$}      & \scriptsize{2}                       \\
                         & \scriptsize{${[}1376, 1655{[}$}     & \scriptsize{1}                       \\
                         & \scriptsize{${[}1655, +\infty{[}$} & \scriptsize{0}                       \\ \hdashline
\multirow{5}{*}{\scriptsize{D\_MS}}   & \scriptsize{${[}0, 3780{[}$}         & \scriptsize{4}                       \\
                         & \scriptsize{${[}3780, 4255{[}$}      & \scriptsize{3}                       \\
                         & \scriptsize{${[}4020, 4255{[} $}     & \scriptsize{2}                       \\
                         & \scriptsize{${[}4255, 4490{[}$}      & \scriptsize{1}                       \\
                         & \scriptsize{${[}4490, +\infty{[}$} & \scriptsize{0}                       \\ \hdashline
\multirow{5}{*}{\scriptsize{TTC\_MS}} & \scriptsize{${[}0, 359{[}$}          & \scriptsize{4}                       \\
                         & \scriptsize{${[}359, 394{[}$}        & \scriptsize{3}                       \\
                         & \scriptsize{${[}394, 429{[}$}        & \scriptsize{2}                       \\
                         & \scriptsize{${[}429, 464{[}$}        & \scriptsize{1}                       \\
                         & \scriptsize{${[}464, +\infty{[}$}  & \scriptsize{0}                       \\ \bottomrule
\end{tabular}

\scriptsize{
\begin{tablenotes}
  %\footnotesize   %% If you want them smaller like foot notes
  \item[a] Note: C: Collision occurrence, MD: Minimum relative distance, D\_MS: Distance at maximum relative speed, TTC\_MS: Time to collision at maximum relative speed
\end{tablenotes}}
\end{table} 

        \subsubsection{GA Parameter Tuning}

            The GA employs three operations - selection, crossover, and mutation - each associated with a probability: selection rate ($\mu_s$), crossover rate ($\mu_c$), and mutation rate ($\mu_m$), respectively, such that $\mu_s + \mu_c + \mu_m = 1$. During each generation, while the new population has not yet reached the target size, an operation is selected by randomly sampling a value from a uniform distribution in the range $[0, 1]$ and mapping it to one of the three operations based on their probabilities. If the value falls within $[0, \mu_{s}]$, \textbf{elitism} is applied, selecting the best individual not already chosen from the prior generation, ensuring no repetition \cite{katoch2021review}. If it lies in $[\mu_{s}, (\mu_{s} + \mu_{c})[$, \textbf{single-point crossover} occurs between two parents, producing two new individuals \cite{katoch2021review}. Finally, if the value is in $[(\mu_{s} + \mu_{c}), 1]$, \textbf{random mutation} is applied within a predefined range \cite{katoch2021review}. For the experiments, $\mu_{s} = 0.1$, $\mu_{c} = 0.8$, and $\mu_{m} = 0.1$ were selected based on empirical results. A 10\% mutation rate was adopted to increase variability in the search space while preserving high-quality solutions from earlier generations \cite{grefenstette1986optimization}.

        \subsection{Evalutation Metrics} \label{sub-sec:evaluation_metrics}

            Five evaluation metrics were used to assess the GA performance: Risk Level \textbf{(RL)}, Number of Collisions \textbf{(NC)}, Minimum Distance of all valid (global) scenarios \textbf{(MDG)}, Minimum Distance Excluding Collisions \textbf{(MDEC)}, and Number of Invalids \textbf{(NIS)}. The RL quantifies overall risk, calculated with the same fitness function used in the GA, scoring simulations from \( \{-1\} \cup \{ x \in \mathbb{Z} \mid 0 \leq x \leq 22 \} \). The NC \(\{ x \in \mathbb{Z} \mid 0 \leq x \leq 100 \} \) accounts for the total collisions observed, directly reflecting the scenario criticality. The Minimum Distance (MD) \(\{ x \in \mathbb{Z} \mid x \geq 0 \} \) measures the shortest distance between vehicles during the simulation, with smaller values indicating higher risk. The MD was divided into two subcategories to evaluate the behavior of distances considering (i) set of all simulations (\textit{collisions and non-collisions}) - MDG; (ii) set of simulations in which there is \textit{non-collision} - MDEC. Finally, the NIS \(\{ x \in \mathbb{Z} \mid 0 \leq x \leq 100 \} \) evaluates utilization by counting the simulations that failed validity criteria. Together, these metrics offer a comprehensive view of scenario quality and optimization performance \cite{nascimento2019systematic, kim2022drivefuzz, zhu2023rita, shu2021test, tian2022generating, song2022scenario}.

            The combination of simulation infrastructure, scenario optimization through GA, and structured evaluation metrics constitutes the foundation of the \texttt{CORTEX-AVD} framework. To assess the applicability and performance of this approach, the next section presents a case study involving its deployment in a controlled experimental setting.

\section{Case Study} \label{sec:case_study}
    
\begin{figure}[h]
        \centering
        \includegraphics[scale=0.8]{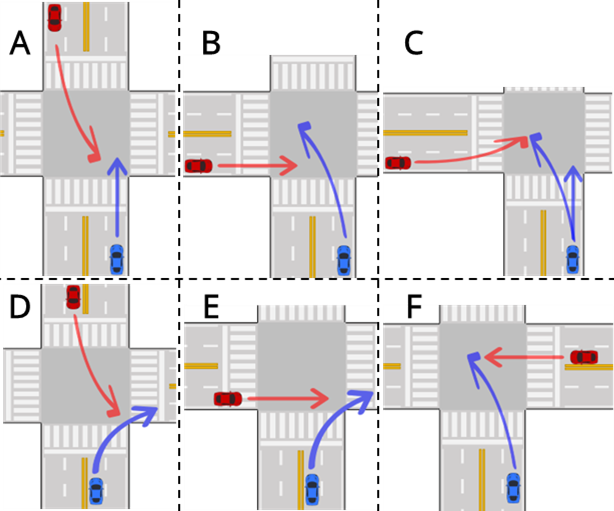}
        % \caption{Set of test scenarios used in the case study according to Table \ref{tab:nhtsa_scenarios}}
        \caption{Illustration of the NHTSA intersection scenarios used in the case study, corresponding to those listed in Table \ref{tab:nhtsa_scenarios}}
        \label{fig:nhtsa_scenarios}
    \end{figure}

    The case study aimed to validate the hypothesis \textbf{H1: simulations generated by GA have a higher likelihood of collision or near-collision occurrences compared to random sampling method}. Intersection scenarios involving two moving vehicles were selected based on the NHTSA’s 2011–2015 light vehicle pre-crash statistics \cite{swanson2019statistics}, using scenarios with high accident incidence \cite{nhtsa_automated_vehicles_safety}. These scenarios, listed in Table \ref{tab:nhtsa_scenarios} and illustrated in Figure \ref{fig:nhtsa_scenarios}, were used to generate relevant data on high-risk events, supported by the module responsible for tuning parameters.

\setlength{\tabcolsep}{1pt}
% \begin{table*}[h]
\begin{table}[h]
\caption{Description of Traffic Scenarios and Vehicle Maneuvers}
\label{tab:nhtsa_scenarios}
\centering
% \begin{tabularx}{\textwidth}{|m{2.5cm}|m{9cm}|m{5cm}|}
% \begin{tabularx}{\textwidth}{|m{2.5cm}|m{2.8cm}|m{2.8cm}|}
\begin{tabularx}{\textwidth}{|m{1cm}|m{0.8cm}|m{6.8cm}|}
 % \toprule
 \cmidrule(){1-3}
 \scriptsize{\textbf{Scenario}} & \scriptsize{\textbf{\#Lanes}} & \scriptsize{\textbf{Description}}\\ 
 % \hline
 \cmidrule(){1-3}
 \scriptsize{\multirow{2}{*}{A}}  & \scriptsize{\multirow{2}{*}{2 x 2}} & \scriptsize{\textbf{Vehicle 1} performs a \textbf{\underline{crossing}}} \\

&   &   \scriptsize{\textbf{Vehicle 2} (same lane, opposite direction) performs a \textbf{\underline{left turn}}}               \\

 % \hline
 \cmidrule(){1-3}
 \scriptsize{\multirow{2}{*}{B}}     & \scriptsize{\multirow{2}{*}{2 x 2}}                   & \scriptsize{\textbf{Vehicle 1} performs a \textbf{\underline{left turn}}} \\ 

&   &   \scriptsize{\textbf{Vehicle 2} (perpendicular lane) performs a \textbf{\underline{crossing}}}                         \\ 
 % \hline
 \cmidrule(){1-3}

\scriptsize{\multirow{2}{*}{C}}  &

 \scriptsize{\multirow{2}{*}{2 x 2}}                   & \scriptsize{\textbf{Vehicle 1} performs a \textbf{\underline{crossing}} or \textbf{\underline{left turn}}} \\

&   &   \scriptsize{\textbf{Vehicle 2} (perpendicular lane) performs a \textbf{\underline{left turn}}} \\

\cmidrule(){1-3}

\scriptsize{\multirow{2}{*}{D}}                & \scriptsize{\multirow{2}{*}{2 x 2}}                   & \scriptsize{\textbf{Vehicle 1} performs a \textbf{\underline{right turn}}} \\

&   &   \scriptsize{\textbf{Vehicle 2} (same lane, opposite direction) performs a \textbf{\underline{left turn}}}        \\

\cmidrule(){1-3}

\scriptsize{\multirow{2}{*}{E}}                & \scriptsize{\multirow{2}{*}{2 x 2}}                   & \scriptsize{\textbf{Vehicle 1} performs a \textbf{\underline{right turn}}} \\
&   &   \scriptsize{\textbf{Vehicle 2} (perpendicular lane) \textbf{\underline{crosses}} in the same direction as A}\\
\cmidrule(){1-3}

\scriptsize{\multirow{2}{*}{F}}                & \scriptsize{\multirow{2}{*}{3}}                       & \scriptsize{\textbf{Vehicle 1} performs a \textbf{\underline{left turn}}} \\
&   &   \scriptsize{\textbf{Vehicle 2} (perpendicular lane) \textbf{\underline{crosses}} in the same direction as A} \\

 % \hline
 \cmidrule(){1-3}
\end{tabularx}
% \end{table*}
\end{table}

\begin{figure}[h]
        \centering
        \includegraphics[scale=0.35]{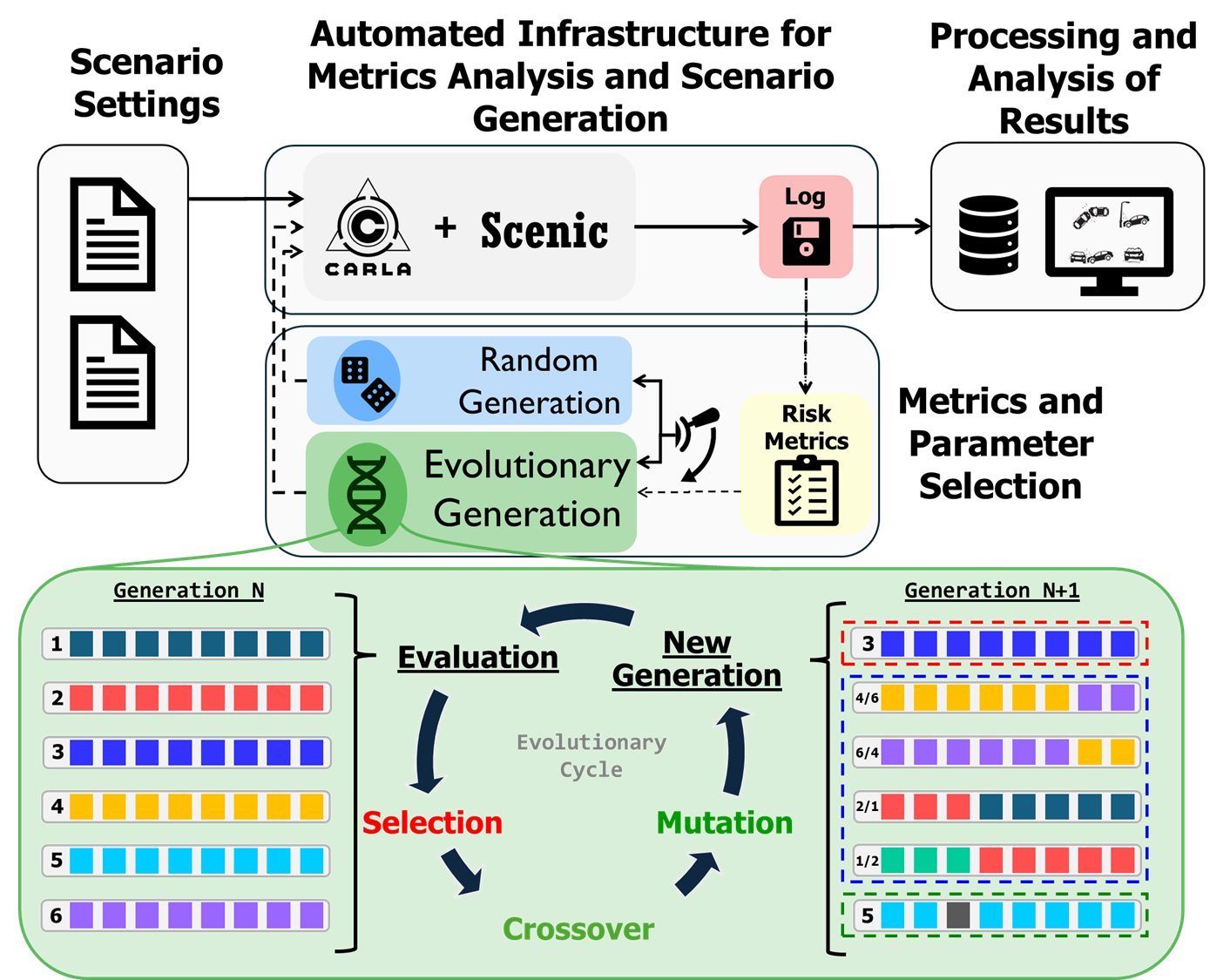}
        \caption{Case study infrastructure}
        \label{fig:case_study_infrastructure}
    \end{figure}

    To test H1, 36,000 crossover simulations were run across all test scenarios - 18,000 generated randomly and 18,000 using GA - requiring around 135 hours of continuous experiment execution. The complete infrastructure is shown in Figure \ref{fig:case_study_infrastructure}. The GA was configured to run for 30 generations, with each generation composed by 100 distinct individuals. These values were selected based on preliminary experiments that indicated they provided a good balance between performance and computational cost. A "\textit{generation}" refers to one iteration of the GA cycle, during which a new population of individuals is created through selection, crossover, and mutation.
    
    To ensure a standardized comparison between the GA and random approaches, the same number of simulations was used for each. Specifically, for each generation, 100 scenarios were generated by the GA and 100 scenarios were produced by sampling random parameters, ensuring symmetry in simulation structure and volume. This setup ensured that both methods operated under equivalent conditions for comparative analysis.

\section{Results} \label{sec:results}

\begin{figure*}[h]
        \centering
        \includegraphics[scale=0.087]{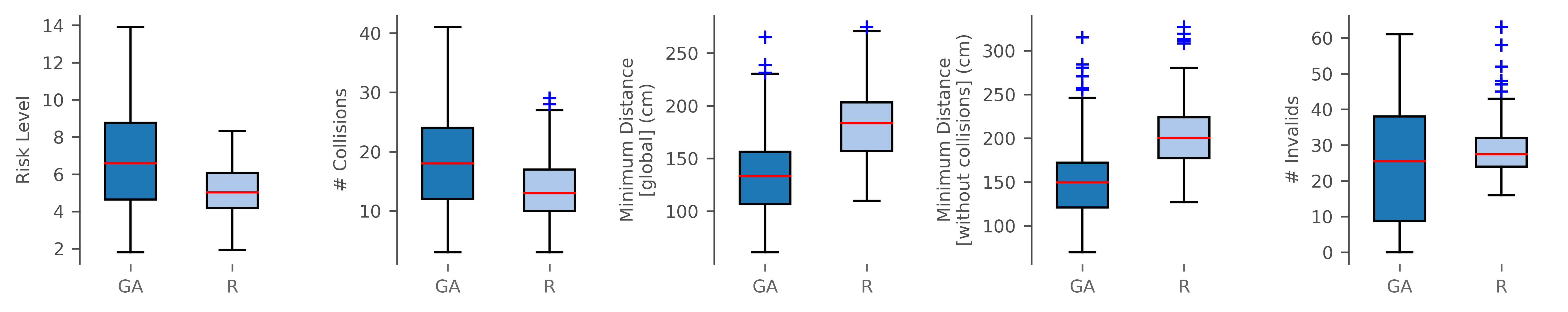}
        \caption{Boxplot comparing five evaluation metrics of the average results of all scenarios.}
        \label{fig:boxplot_consolidado}
    \end{figure*}

    Figure \ref{fig:boxplot_consolidado} shows boxplots with the average results across the six tested scenarios, highlighting GA's advantages in all evaluation metrics. The results, analysis, and discussion are then grouped according to the four performance metrics: RL, NC, MDG, MDEC, and NIS. To facilitate the analysis, the style (solid, dashed curves) and color patterns are shared among the Figures \ref{fig:result_risk} to \ref{fig:invalidos}. In these figures, a Savitzky-Golay filter was applied to smooth the noisy curves (12 per plot), improving readability and highlighting average trends across generations \cite{schafer2011savitzky}.

    \begin{figure*}[h]
        \centering
        \includegraphics[scale=0.08]{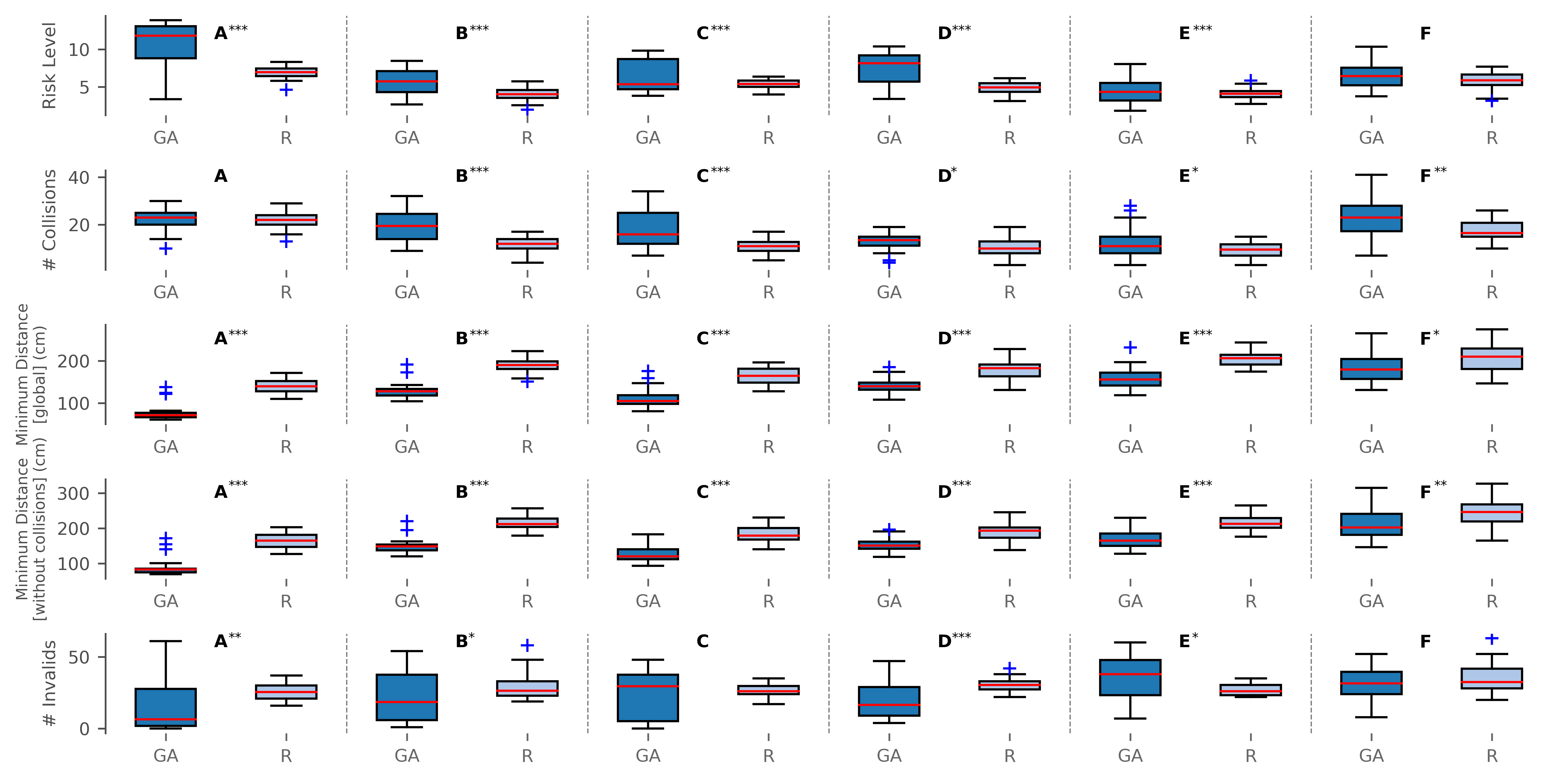}
        \caption{Boxplot comparing five evaluation metrics of the average results for each scenario.}
        \label{fig:boxplot_cenarios}
        \scriptsize{Note: The symbols used in this table indicate the level of statistical significance. The symbol "***" indicates a \textit{p-value} less than or equal to 0.001, "**" denotes a \textit{p-value} less than or equal to 0.01, "*" denotes a \textit{p-value} less than or equal to 0.05.}
    \end{figure*}

    \subsection{Risk Level}

\begin{figure}[h]
            \centering
            \includegraphics[scale=0.092]{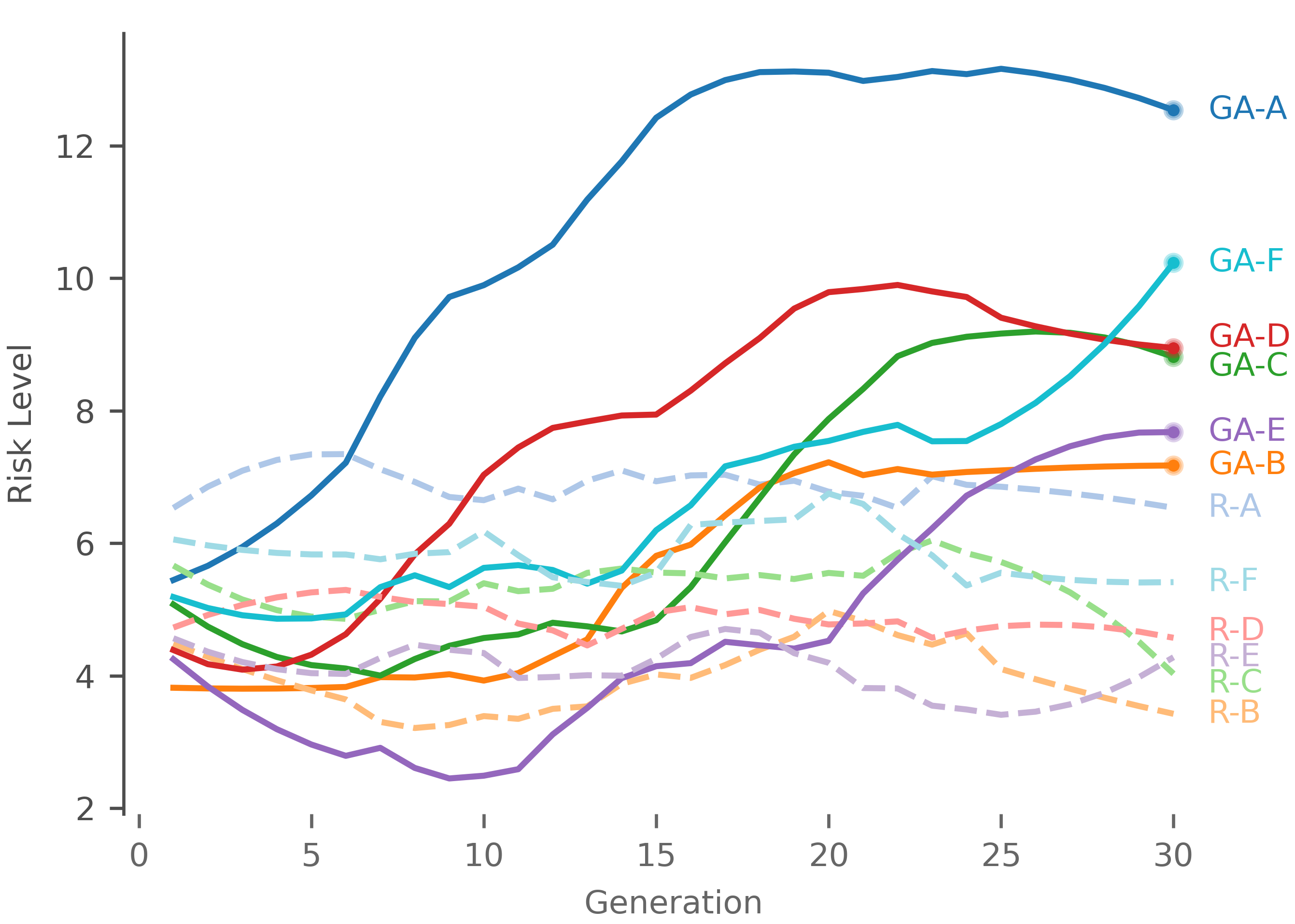}
            \caption{Comparison of RLs in the six scenarios.}
            \label{fig:result_risk}
        \end{figure}

        As presented in Figure \ref{fig:result_risk}, GA shows non-monotonic behavior, especially in the early generations, with oscillations in C and E. After these initial phases, GA consistently outperforms the Random approach, particularly from the middle to final generations, where the difference becomes statically significant in most scenarios. Convergence occur earlier in scenarios A, B, C, and D, while E and F show an improvement in performance in the last generations, suggesting GA parameter optimization could improve performance.

        The boxplots in the first row of Figure \ref{fig:boxplot_cenarios} show that, for each scenario, GA is the most effective technique to improve RL compared to Random heuristic. The following presents the average RL of each experiment with the following pattern \(\{ Scenario \in [A, B, C, D, E, F]: \textbf{GA} - Random \} \) -  \(\{A: \textbf{13.3} - 9.7 \mid B: \textbf{7.5} - 6.1 \mid C: \textbf{8.8} - 7.6 \mid D: \textbf{9.8} - 7.5 \mid E: \textbf{7.7} - 6.0 \mid F: \textbf{10.1} - 9.7\} \). Similarly, Figure \ref{fig:boxplot_consolidado} shows that GA increases the overall average RL from 7.75 to 9.54, an 18.7\% relative increase, indicating that optimization improves performance.

    \subsection{Number of Collisions}

\begin{figure}[h]
            \centering
            \includegraphics[scale=0.092]{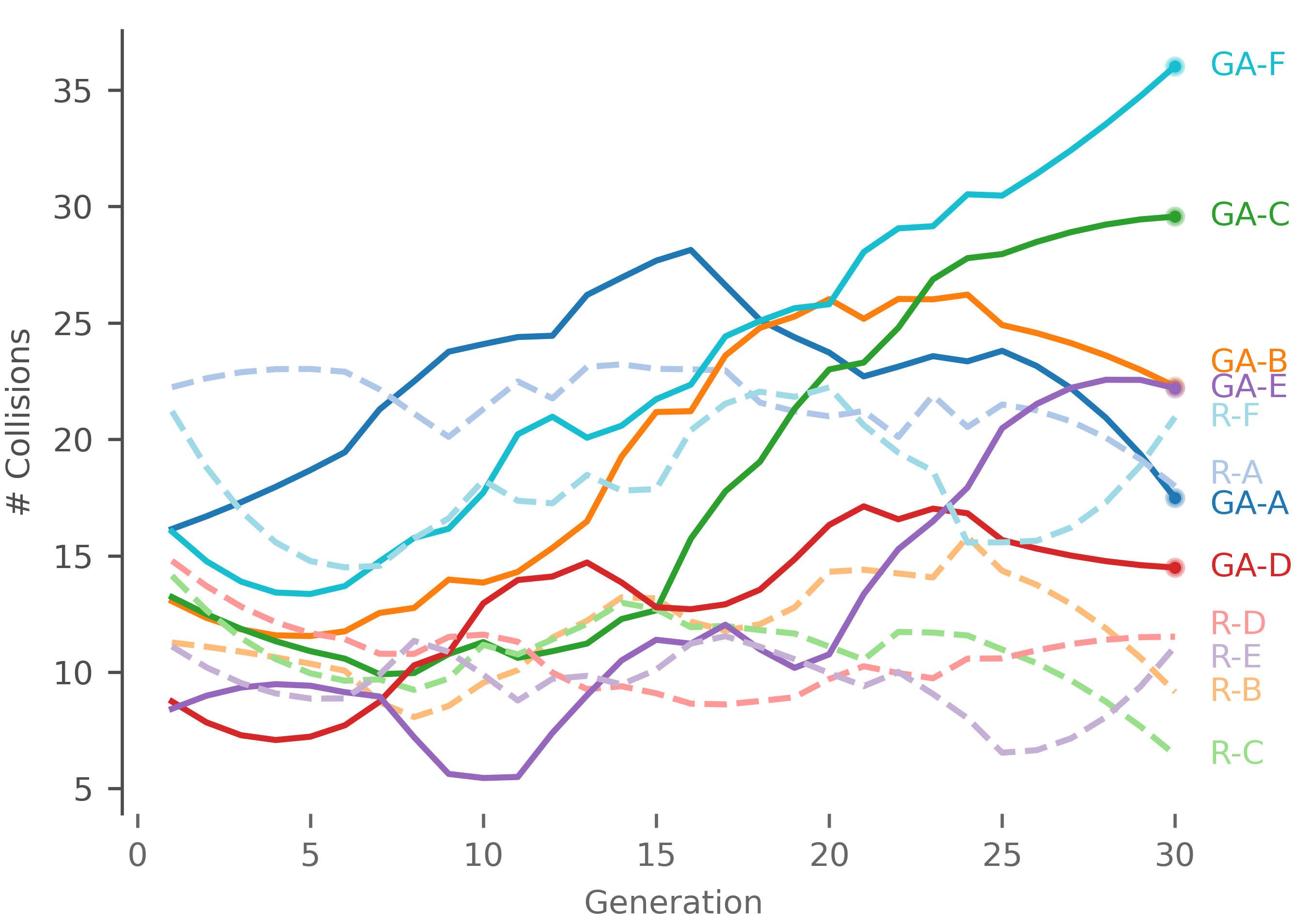}
            \caption{Comparison of the NCs in the six scenarios.}
            \label{fig:result_collisions}
        \end{figure}

        Collision analysis reveals a distinct pattern, which in Random experiments, the NCs present a stable tendency between generations, whereas GA shows fewer collisions in early generations, followed by increased collisions later, as shown in Figure \ref{fig:result_collisions}. Notably, scenarios A (\textbf{GA: 673} | R: 651), D (\textbf{GA: 386} | R: 322), and E (\textbf{GA: 377} | R: 284) did not show significant differences in total collisions between methods. In contrast, scenarios B (\textbf{GA: 577} | R: 355), C (\textbf{GA: 544} | R: 327), and F (\textbf{GA: 692} | R: 543) show a significant higher NCs compared to Random generation, suggesting that the performance of GA may vary significantly depending on the specific scenario. However, breaking down collisions across generations [21-30] shows that the final stages of GA consistently yield higher risk scenarios A (\textbf{GA: 220} | R: 204), B (\textbf{GA: 249} | R: 133), C (\textbf{GA: 273} | R: 101), D (\textbf{GA: 158} | R: 109), E (\textbf{GA: 195} | R: 82), and F (\textbf{GA: 312} | R: 174), aligning with the hypothesis. The correlation between collisions and risk level is high, as collisions add a substantial value to the risk score (45\% of score value), reinforcing GA's effectiveness in generating CC. In most scenarios, GA tends to produce more collisions as generations progress, diverging from the Random approach, which may indicate that GA prioritizes CC, leading to more collisions while exploring extreme conditions. The tendencies described before are further reinforced by the boxplots of Figure \ref{fig:boxplot_consolidado} and the second row of Figure \ref{fig:boxplot_cenarios}.

    \subsection{Minimum Distances}

\begin{figure}[h]
            \centering
            \includegraphics[scale=0.089]{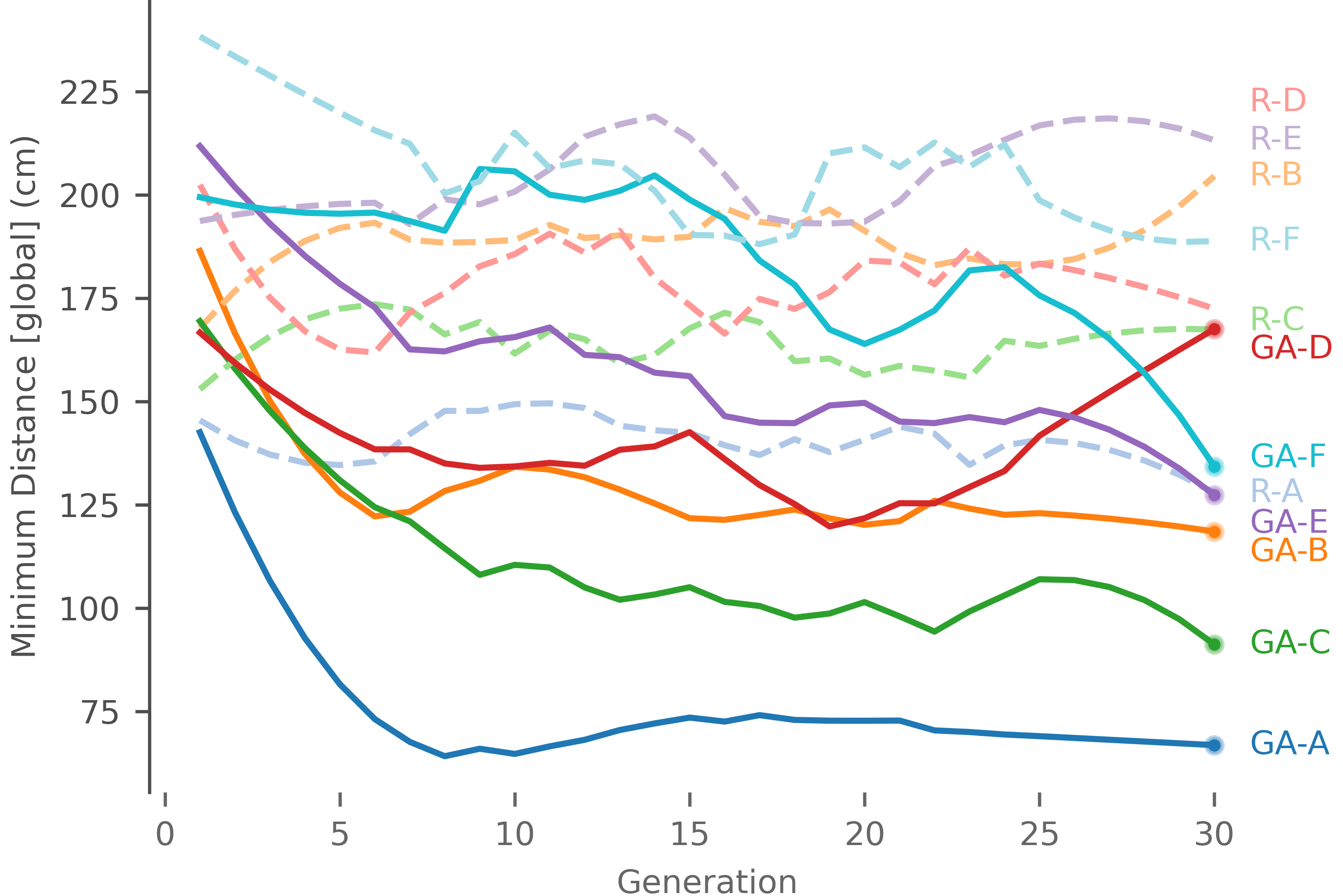}
            \caption{Comparison of MD considering all valid simulations in the six scenarios.}
            \label{fig:dist_min_global}
        \end{figure}

\begin{figure}[h]
            \centering
            \includegraphics[scale=0.089]{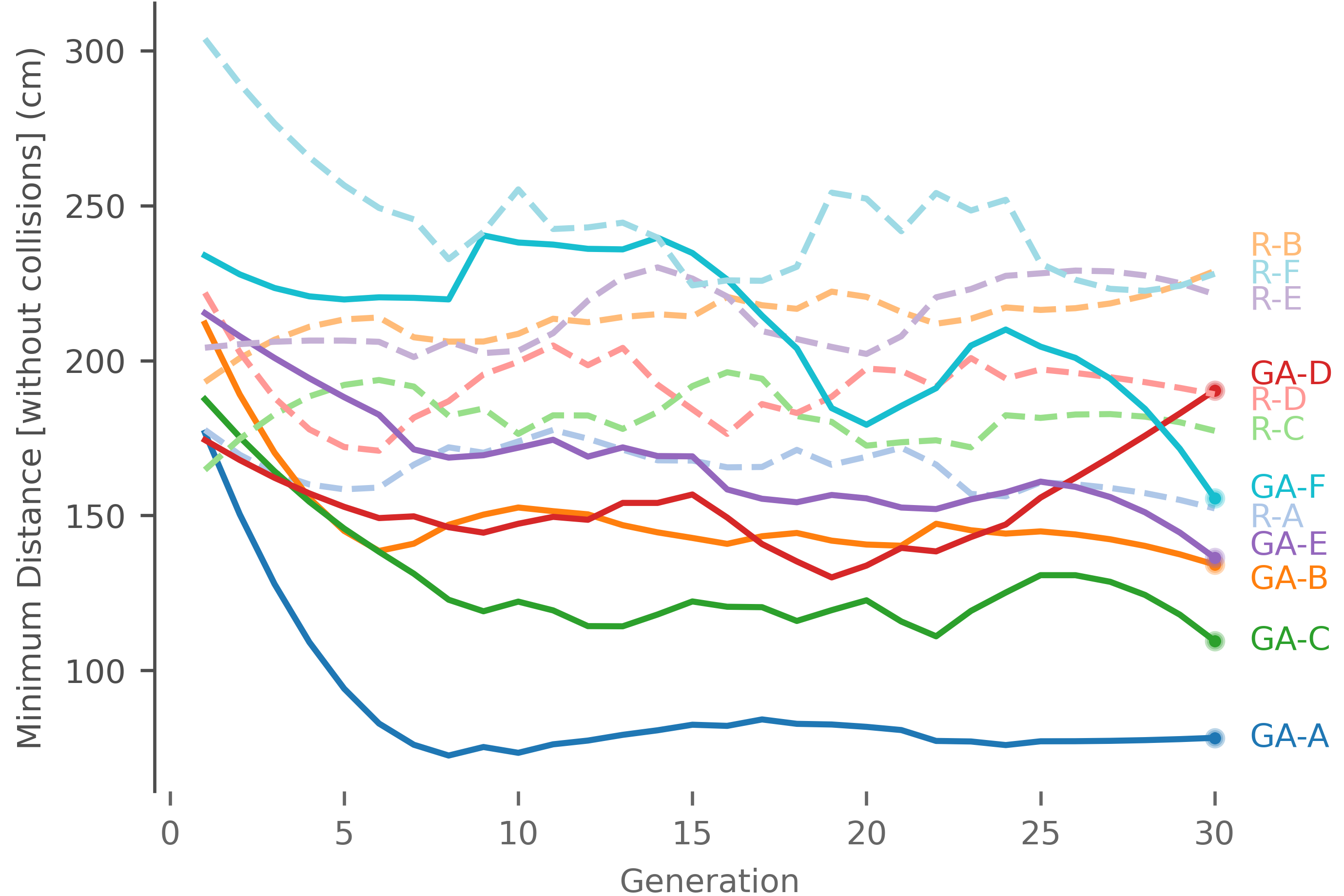}
            \caption{Comparison of MD excluding simulations with collisions in the six scenarios.}
            \label{fig:dist_min_no_collision}
        \end{figure}

        The MD metric evaluates the closest proximity between vehicles to assess risk during simulations, with the level of risk depending on the specific actions and context of the vehicles involved. In normal traffic, vehicles may approach each other without posing significant risk, but in these experiments, which forces confrontations in conflict zones during crossing and turning actions, the MD reflects the aggressiveness of their approach \cite{bhatt2022driver}. Figures \ref{fig:dist_min_global} and \ref{fig:dist_min_no_collision} show the average minimum approach distances (in centimeters), with the first considering all valid scenarios (MDG) and the second excluding simulations that resulted in collisions (MDEC) to focus on near-accident situations. As seen with previous metrics, GA scenarios exhibit a pattern of increasing risk, with MDs decreasing over generations, while Random approach averages remain near-uniform across generations. The trends are well represented by Figure \ref{fig:boxplot_consolidado} and the third and fourth rows of Figure \ref{fig:boxplot_cenarios}, highlighting the superiority of GA in all scenarios, despite the regression of MD in scenario D in the final generations. Notably, in the last 10 generations [21-30], MDG showed GA’s superiority across all scenarios: A (\textbf{GA: 69.4} | R: 138.9), B (\textbf{GA: 121.5} | R: 187.5), C (\textbf{GA: 101.4} | R: 162.5), D (\textbf{GA: 144.6} | R: 180.8), E (\textbf{GA: 143.9} | R: 213.6), and F (\textbf{GA: 166.8} | R: 200.7). In MDEC, even with collisions removed, showed similar results, though the average distances were slightly higher for all scenarios: A (\textbf{GA: 77.9} | R: 161.2), B (\textbf{GA: 141.8} | R: 217.6), C (\textbf{GA: 122.6} | R: 178.2), D (\textbf{GA: 160.9} | R: 195.7), E (\textbf{GA: 154.7} | R: 224.6), and F (\textbf{GA: 191.5} | R: 236.1). Thus, GA consistently generated riskier scenarios with decreasing MDs, even in MDEC, reinforcing its ability to explore near-accident conditions more effectively than Random method.

        \subsection{Invalid Scenarios}

\begin{figure}[h]
                \centering
                \includegraphics[scale=0.094]{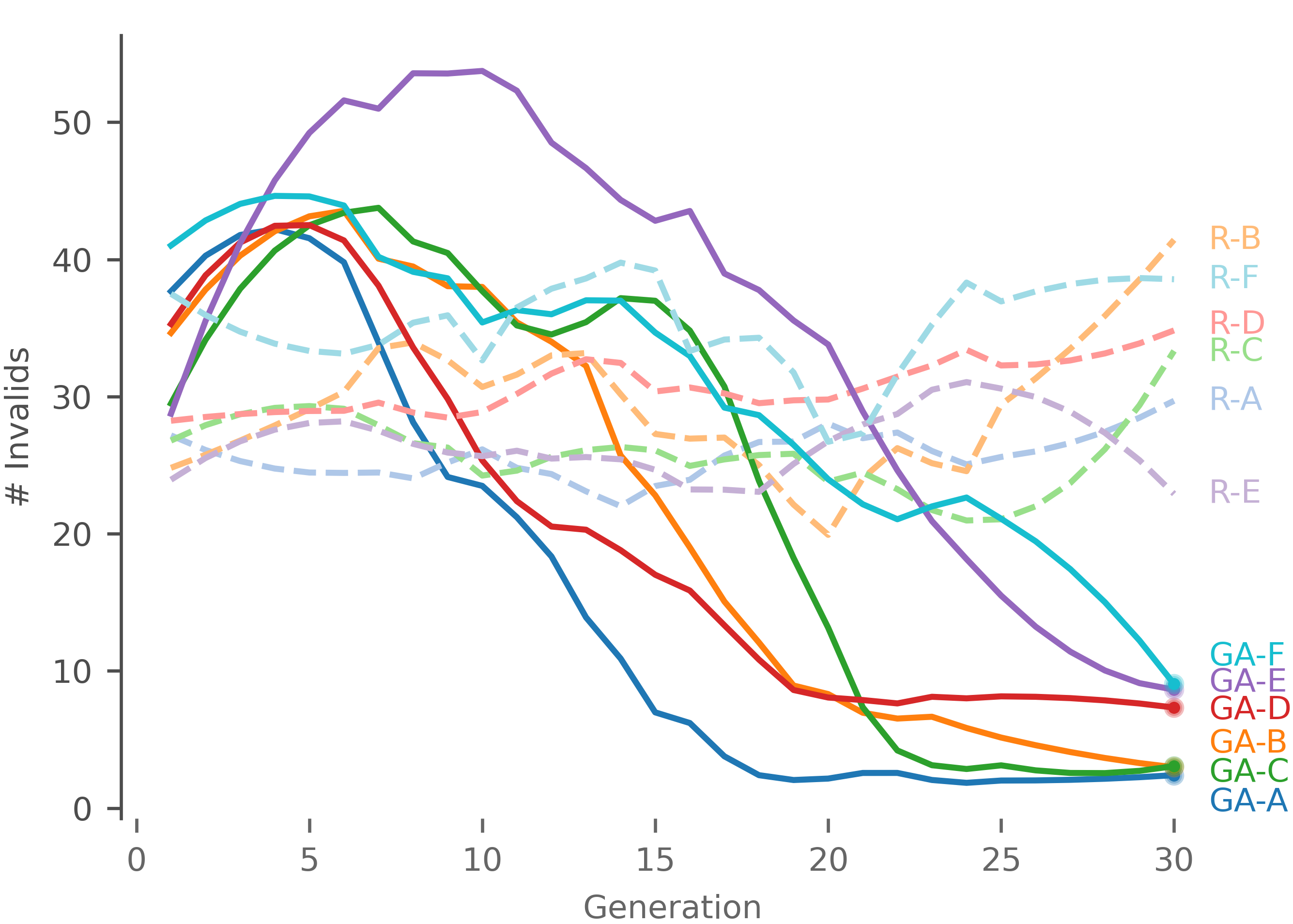}
                \caption{Comparison of the NISs in the six scenarios.}
                \label{fig:invalidos}
            \end{figure}

            Scenic sometimes generates scenario descriptions that fail to produce executable simulations. To assess the GA's computational efficiency, the NIS is counted, where a lower NIS in a fixed set of tests indicates greater effectiveness. Figure \ref{fig:invalidos} shows the NIS simulations per scenario, similar to collision analysis. The Random heuristic exhibits a near-uniform distribution across generations and in the global average, whereas GA shows a non-monotonic trend where NIS increases in the initial generations before decreasing as generations progress. Notably, although the total NIS counts for scenarios C (\textbf{GA: 725} | R: 778), E (\textbf{GA: 1047} | R: 798), and F (\textbf{GA: 922} | R: 1059) were less favorable, scenarios A (\textbf{GA: 468} | R: 771), B (\textbf{GA: 665} | R: 885), and D (\textbf{GA: 604} | R: 924) demonstrated significant improvement in NIS reduction, as supported by the fifth row of Figure \ref{fig:boxplot_cenarios}. The most remarkable improvements occurred in the last 10 generations [21-30], where reductions were observed across all scenarios, with A (\textbf{GA: 20} | R: 267), B (\textbf{GA: 50} | R: 312), C (\textbf{GA: 36} | R: 249), D (\textbf{GA: 77} | R: 325), E (\textbf{GA: 159} | R: 286), and F (\textbf{GA: 187} | R: 371). The improvement rates ranged from 44\% (scenario E) to 93\% (scenario A). Finally, as shown in Figure \ref{fig:boxplot_consolidado}, despite the larger distribution, GA shows a clear trend to reduce NIS over successive generations.

\section{Discussion}\label{sec:discussion}

    In the case study, GA consistently outperformed the Random approach across five key metrics, providing strong support for H1. Risk Level (RL) increased in all scenarios, with gains ranging from 4\% (scenario F) to 37\% (scenario A), averaging a 23\% improvement. GA also collected 31\% more collisions (3,249 vs. 2,482), highlighting its ability to explore riskier situations. The average Global Minimum Distance (MDG) between vehicles dropped from 180.6cm to 133.2cm with GA, a 26\% decrease, while Minimum Distance Excluding Collision (MDEC) decreased from 202.2cm to 149.6cm, a 26\% decrease, indicating a growing driving risk due to closer vehicle proximity. Notably, the MDEC suggests that GA not only generates collisions but also increases near-collision cases, enriching data for AV evaluation. This contribution is significant, as near-collisions, often neglected in peer studies, provide crucial insights for AV evaluation and risk assessment. Additionally, GA reduced Number of Invalid Scenarios (NIS) by 15\%, enhancing the utilization rate (UR) of the CC generator infrastructure. Notably, GA's data generation is economical and energy-efficient: a 100k simulation log requires about 10GB, whereas a database of sensor data (e.g. camera, lidar) from these logs would require tens to hundreds of terabytes, depending on sensor resolution quality. This implies improved efficiency in generating test cases at a fixed computational processing and storage cost.

    Despite growing interest in heuristic-based CC generation, standardized benchmarks remain scarce. For that reason, an effort was made to organize the literature using a set of common metrics (Number of Test Cases, Time, Number of Collisions (NC), Number of Valid Scenarios), used for GA efficiency evaluation (Table \ref{tab:study_comparison}). However, no metrics related to near-collisions were listed, as none of these studies explicitly collected this information, making comparisons impossible. Variations across studies often reflect differing focuses, such as topology generation \cite{tian2022mosat, gambi2019automatically, tang2021systematic} or risk-related metrics like time \cite{kluck2019genetic, kluck2023empirical, kluck2019performance}, distances \cite{gambi2019automatically, tang2021systematic, ebadi2021efficient, kaufmann2021critical}, and collision rates \cite{tian2022mosat, ebadi2021efficient, kaufmann2021critical}. While recent studies underscore GA's value for CC generation, most experiments involve only a few thousand simulations and focus on limited metrics \cite{kluck2019performance, tian2022mosat, gambi2019automatically, birchler2023single, langford2019applying, tang2021systematic, ebadi2021efficient, kluck2019genetic, kaufmann2021critical, zhou2023specification}. RL results are also highly dependent on GA modeling and simulation frameworks, making cross-study comparisons challenging.

    \renewcommand{\dashlinedash}{0.5pt} % Dash length
\renewcommand{\dashlinegap}{1.5pt} % Gap length

\setlength{\tabcolsep}{1pt}
% \begin{table*}[h]
\begin{table}[h]
\caption{Comparison of studies}
\label{tab:study_comparison}
\centering
% \begin{tabularx}{\textwidth}{|m{2.5cm}|m{9cm}|m{5cm}|}
% \begin{tabularx}{\textwidth}{|m{2.5cm}|m{2.8cm}|m{2.8cm}|}
\scriptsize

\begin{threeparttable}
\begin{tabularx}{\textwidth}{|m{1.0cm}|m{1cm}|m{1.1cm}|m{1.7cm}|m{1.8cm}|m{1.5cm}|}
% \begin{tabularx}{\textwidth}{|x|x|x|x|x|x|}
 % \toprule
 \cmidrule(){1-6}
 \textbf{Study} & \textbf{\#Test Cases} & \textbf{Time{\normalsize \tnote{*}} (s)} & \textbf{\#Collisions} & \textbf{\#Valid Scenarios} & \textbf{Metrics} \\ 
 % \hline
 \cmidrule(){1-6}
 \cite{kluck2019performance} &
  9,135 &
  NS {\normalsize \tnote{a}} &
  UOM {\normalsize \tnote{b}}&
  MNC {\normalsize \tnote{c}} &
  TTC {\normalsize \tnote{d}} \\

  \hdashline
  
 % \hline
 % \cmidrule(){1-6}
 
 \cite{tian2022mosat} &
  50,000** &
  % \begin{tabular}[c]{@{}c@{}}24 (MOSAT)\\ 55 (AV-Fuzzer)\end{tabular} &
  24-55 &
  1.1\%-2.85\% &
  UOM &
  % \begin{tabular}[c]{@{}c@{}}METTC, DFP,\\ VOA, AEDF\end{tabular} \\
  METTC, DFP, VOA, AEDF {\normalsize \tnote{f}} \\

  \hdashline
 % \hline
 % \cmidrule(){1-6}

\cite{gambi2019automatically} &
  TB {\normalsize \tnote{e}} &
  % \begin{tabular}[c]{@{}c@{}}41.1 (single-road)\\ 152.6 (multi-road)\end{tabular} &
  41.1-152.6 &
  NS &
  % \begin{tabular}[c]{@{}c@{}}34\% (single-road)\\ 39\% (multi-road)\end{tabular} &
  34\% to 39\% &
  Distance \\

  \hdashline

 % \cmidrule(){1-6}

 \cite{langford2019applying} &
  4,989 &
  UOM &
  NS &
  NS &
  Distance \\

    \hdashline
 % \cmidrule(){1-6}

 \cite{tang2021systematic} &
  10,240 &
  60 &
  NS &
  UOM &
  Distance \\

\hdashline
 % \cmidrule(){1-6}

 \cite{ebadi2021efficient} &
  200 &
  10*** &
  % \begin{tabular}[c]{@{}l@{}}GA: 15 (7.5\%)\\R: 8 (4\%)\end{tabular} &
  \begin{tabular}[c]{@{}l@{}}{\fontsize{6}{10}\selectfont \textbf{GA}: 15 (7.5\%)}\\{\fontsize{6}{10}\selectfont \textbf{R}: 8 (4\%)}\end{tabular} &
  MNC &
  Distance \\

  \hdashline

 % \cmidrule(){1-6}

 \cite{kluck2019genetic} &
  2,342 &
  NS &
  72 (3.1\%) &
  % \begin{tabular}[c]{@{}l@{}}GA: 878 (25\%)\\ R: 468 (40\%)\end{tabular} &
  \begin{tabular}[c]{@{}l@{}}{\fontsize{6}{10}\selectfont \textbf{GA}: 878 (25\%)}\\{\fontsize{6}{10}\selectfont \textbf{R}: 468 (40\%)}\end{tabular} &
  TTC \\

  \hdashline

 % \cmidrule(){1-6}

 \cite{kaufmann2021critical} &
  1,800 &
  10*** &
  233 (12.9\%) &
  UOM &
  Distance \\

  \hdashline

 % \cmidrule(){1-6}

 \cite{zhou2023specification} &
  40,000 &
  56 &
  % \begin{tabular}[c]{@{}l@{}}GA: 2,666 (13.3\%)\\ R: 747 (3.7\%)\end{tabular} &
  \begin{tabular}[c]{@{}l@{}}{\fontsize{6}{10}\selectfont \textbf{GA}: 2,666 (13.3\%)}\\{\fontsize{6}{10}\selectfont \textbf{R}: 747 (3.7\%)}\end{tabular} &
  MNC &
  NS \\

  % \hdashline[0.1pt]
  \hdashline

 % \cmidrule(){1-6}

 \cite{kluck2023empirical} &
  83,726 &
  NS &
  % \begin{tabular}[c]{@{}l@{}}GA: 5,946 (30.7\%)\\ R: 729 (3.7\%)\\ CT: 3,862 (8.6\%)\end{tabular} &
  \begin{tabular}[c]{@{}l@{}}{\fontsize{6}{10}\selectfont \textbf{GA}: 5,946 (30.7\%)}\\{\fontsize{6}{10}\selectfont \textbf{R}: 729 (3.7\%)}\\ {\fontsize{6}{10}\selectfont \textbf{CT}: 3,862 (8.6\%)}\end{tabular} &
  MNC &
  TTC \\

 \cmidrule(){1-6}

 \scriptsize{\textbf{\texttt{CORTEX}}} &
  36,000 &
  13.5 &
  % \begin{tabular}[c]{@{}l@{}}GA: 3,249 (18.1\%)\\ R: 2,482 (13.8\%)\end{tabular} &
  \begin{tabular}[c]{@{}l@{}}{\fontsize{6}{10}\selectfont \textbf{GA: 3,249 (18.1\%)}}\\{\fontsize{6}{10}\selectfont \textbf{R: 2,482 (13.8\%)}}\end{tabular} &
  % \begin{tabular}[c]{@{}l@{}}GA: 13,569 (75.4\%)\\ R: 12,785 (71\%)\end{tabular} &
  \begin{tabular}[c]{@{}l@{}}{\fontsize{6}{10}\selectfont \textbf{GA: 13,569 (75.4\%)}}\\{\fontsize{6}{10}\selectfont \textbf{R: 12,785 (71\%)}}\end{tabular} &
  RL, NC, MD \\
 % \hline
 \cmidrule(){1-6}
\end{tabularx}
\smallskip
\scriptsize
\begin{tablenotes}
    \RaggedRight
    \item[]* Cost Time per Simulation; ** Estimated; *** Constrained; a. Not stated;
    \item[]b. Used with other meaning; c. Mentioned but not collected; d. Time-to-Collision;
    \item[]e. Time budget in hours; f. \textbf{METTC}: Minimal Estimation TTC; \textbf{DFP}: Deviation
    \item[]from Planned Route; \textbf{VOA}: Variation rate of Acceleration;
    \item[]\textbf{AEDF}: Average Euclidian Distance for Found safety-violation scenarios
\end{tablenotes}
\end{threeparttable}
% \end{table*}
\end{table}

    Although direct comparisons are difficult, some insights emerged. The test set in this study is larger than seven studies \cite{kluck2019performance, gambi2019automatically, langford2019applying, tang2021systematic, ebadi2021efficient, kluck2019genetic, kaufmann2021critical} and close to two others \cite{tian2022mosat, zhou2023specification}. Kluck et al. (2023) produced more cases, but their dataset combined GA (25\%), Random (25\%), and Combinatorial Testing (50\%), yielding comparable GA and Random outputs \cite{kluck2023empirical}. Excluding studies where the simulation time was constrained to 10 seconds \cite{ebadi2021efficient, kaufmann2021critical}, the computational cost in this study was favorable, averaging 13.5 seconds per simulation — nearly half the time of the fastest test framework \cite{tian2022mosat}. Regarding collision rates, Kluck et al. (2023) reported a collision proportion of 30.7\% , driven by specific front vehicle (34\%) and pedestrian (19.6\%) collisions \cite{kluck2023empirical}. Our GA implementation achieved an 18.1\% collision rate, an 36\% improvement over the next-best study \cite{zhou2023specification}. Moreover, our valid scenario rate reached 75.4\%, twice as high as the second-best framework \cite{gambi2019automatically}. Although GA's potential to create unfeasible scenarios affects UR \cite{tian2022mosat, tang2021systematic, kaufmann2021critical}, some studies neglected to measure this directly \cite{kluck2019performance, ebadi2021efficient, zhou2023specification, kluck2023empirical}, or applied varying definitions \cite{tian2022mosat, tang2021systematic, kaufmann2021critical}, complicating comparisons.

    Although the framework effectively generated CC data, some areas could be potentially improved. Scenic's rejection sampling introduced variability in generating feasible scenarios, especially when restrictive parameter ranges were applied. Slightly relaxing these ranges could increase the number of high-risk simulations, enhancing the overall effectiveness of the approach. GA's performance depended heavily on initial conditions, stopping criteria, and objective functions, which can be potentially fine tuned \cite{kluck2019performance, tian2022mosat, tang2021systematic, ebadi2021efficient, kluck2019genetic, kaufmann2021critical, kluck2023empirical}. Additionally, testing focused on a single map in Carla, limiting exploration of diverse environments and broader risk factors such as overtaking, variable weather conditions, and obstacles, which affects the generalizability of the proposed approach.

\section{Concluding Remarks} \label{sec:conclusions}

    \texttt{CORTEX-AVD}, a novel simulation framework integrating CARLA and Scenic, was developed and evaluated to generate Corner Cases (CC) scenarios for Autonomous Vehicles (AV) and enhance risk scenario generation through parameter selection techniques. Experimental results supported the hypothesis (H1) that Genetic Algorithm (GA) increases the likelihood of generating high-risk scenarios. GA increased the probability of generating CCs, achieving a Risk Level (RL) gain between 4\% and 37\%, with an average improvement of 23\%, while also reducing the Number of Invalid Scenarios (NIS) by 25\%. Additionally, GA improved the Minimum Distance (MD) between vehicles, increasing near-collision likelihood and enriching datasets with riskier scenarios for AV testing. However, GA exhibited a tendency to converge to local maxima, potentially limiting scenario diversity. Thus, future work should improve the balance between exploration and exploitation, explore alternative optimization algorithms, refine parameter selection, and expand scenario complexity to incorporate diverse risk factors such as weather, pedestrians, and road obstacles. These advancements could improve the understanding of AV performance in CC situations, opening new paths for stronger risk assessment frameworks in simulated environments.

\section*{Acknowledgments}
% This should be a simple paragraph before the References to thank those individuals and institutions who have supported your work on this article.
The Article Processing Charge (APC) for the publication of this research was funded by the Coordenaçãao de Aperfeiçoamento de Pessoal de Nível Superior-Brasil (CAPES). Gabriel Kenji Godoy Shimanuki was supported by Itaú Unibanco S.A., through the Itaú Scholarships Program (PBI), linked to the Data Science Center at the Polytechnic School of the University of São Paulo.

% This work was supported by Itaú Unibanco S.A., through the Itaú Scholarships Program (PBI), linked to the Data Science Center at the Polytechnic School of the University of São Paulo.

\bibliographystyle{IEEEtran}
\bibliography{references}

% \newpage

 

% \end{IEEEbiography}



\vfill

\end{document}